\documentclass{article}
\usepackage{amsmath, amssymb, amsthm, bbm, hyperref, enumitem, booktabs, graphicx, tablefootnote, multirow}
\usepackage{doi}
\usepackage{arxiv}
\usepackage[top=3cm, bottom=3cm, right=3cm, left=3cm]{geometry}
\usepackage{nicematrix} 
\usepackage{tikz}
\usepackage{subcaption}
\usepackage{siunitx}
\usepackage[round]{natbib}
\usepackage{xcolor}
\hypersetup{
  colorlinks   = true, 
  urlcolor     = blue, 
  linkcolor    = red, 
  citecolor   = blue
}

\newcommand{\x}{\mathbf{x}}

\newcommand{\RMSEbar}{\overline{\mathrm{RMSE}}}
\newcommand{\cellstack}[2]{\begin{tabular}{@{}c@{}}#1\\[-3pt]#2\end{tabular}}

\newtheorem*{corollary*}{Corollary}

\providecommand{\keywords}[1]
{
  \small	
  \textbf{\textit{Keywords---}} #1
}

\defcitealias{UN2023manual}{United Nations}
\defcitealias{UN2025model}{2025}

\title{Accurate predictions of properties of energetic materials with tabular foundation models}
\author{\phantom{whitespace} Vagn Krister Nes Glomsrud$^1$ \and Mathias Bynke$^1$\phantom{whitespace} \and Lars Henry Berge Olsen$^2$ \and Dennis Christensen$^1$}
\date{\small $^1$Norwegian Defence Research Establishment (FFI) \\ \small $^2$Norwegian Computing Center (NR)}

\begin{document}

\maketitle

\begin{abstract}
    We use TabPFN, a recently developed tabular foundation model, to predict three properties of energetic materials: impact sensitivity, decomposition temperature and crystal density. Without any hyperparameter tuning, TabPFN outperforms the best published models for all three properties. It also substantially outperforms a wide range of established machine-learning models, including tuned gradient boosting (XGBoost), random forests and Gaussian processes. This holds both when the models are given hand-crafted molecular descriptors and when they are given learned representations from Chemprop's message-passing neural network. In particular, for all three properties, TabPFN applied to Chemprop's own embedding is more accurate than Chemprop's predictions. TabPFN is also among the fastest models we consider, more than an order of magnitude faster than XGBoost. As TabPFN predicts the entire distribution of the target rather than just a single value, it offers built-in uncertainty quantification, which we compare with the empirical distributions of compounds with repeated measurements. Finally, we interpret the predictions using Shapley values, which rank the features by their importance.
\end{abstract}


\section{Introduction}
Accurate assessment of the properties of energetic materials is crucial for ensuring safe production, transportation, handling, usage, storage and destruction of explosives. Unfortunately, some crucial properties require lengthy and costly experimental procedures to quantify properly. Due to this challenge, there has been a surge of interest in predicting properties of energetic materials using machine-learning methods \citep{zang2023prediction}. Some recent examples include prediction of impact sensitivity \citep{lansford2022building, marrs2023chemical, altayeb2026new, hutchings2026predicting, xu2026deep}, crystal density \citep{nguyen2021predicting, yang2021density, wespiser2023application, davis2024machine, pandey2024predicting}, decomposition temperature \citep{wu2023machine, pandey2024predicting}, enthalpy of formation \citep{langer2025prediction} and genotoxicity \citep{fournier2023in}. Furthermore, with the aid of tools from explainable artificial intelligence (XAI), these predictive models can provide new chemical insights into energetic materials.

The task of predicting properties of energetic materials with machine learning is especially challenging, for several reasons. First, measuring the properties mentioned above requires hazardous and sometimes expensive experiments, and many laboratories (both national and private) have restrictive data sharing policies. The result is that there are few openly available high-quality datasets of even moderate size. Second, the chemical descriptors of an energetic material, from which the desired property is to be predicted, are highly heterogeneous in nature. Some are integer-valued (like atomic counts), some are real-valued (like oxygen balance and heat of explosion), and others are categorical (like an identifier of the most sensitive functional group). Third, scientists often employ different definitions of the target to be predicted, which limits the possibility of aggregating different datasets (see e.g.~\citet{christensen2026stop} on sensitivity). A partial solution to this limitation is to employ transfer learning by training the predictive model on a separate larger dataset with pre-computed chemical properties that are physically related to the prediction target \citep{lansford2022building, hutchings2026predicting}.

All the above issues contribute to explaining why deep learning models struggle with predicting properties of energetic materials. Indeed, for moderately sized tabular datasets (i.e.~data in rows and columns) with heterogeneous features, tree-based models such as random forests and XGBoost (which are not based on deep learning) have typically outperformed deep neural networks \citep{grinsztajn2022why}. When neural networks are used for predicting properties of energetic materials, they typically only contain a few hidden layers \citep{yang2021density, lansford2022building, langer2025prediction, altayeb2026new, hutchings2026predicting, xu2026deep}. Moreover, they do not leverage modern developments such as in-context learning or transformer architectures, which have driven the remarkable performance of large language models.

Recently, \citet{hollmann2025accurate} introduced their tabular prior-data fitted network, TabPFN. This is a general tabular foundation model which can quickly perform highly accurate regression and classification on virtually any moderately sized generic tabular dataset. Unlike traditional machine-learning models, TabPFN is not directly trained to predict the target of interest. Instead, it has been pre-trained on a large number of purely synthetic datasets, and makes its predictions for the application in question using in-context learning, where the application-specific training data serve as context. TabPFN has been demonstrated to outperform state-of-the-art models on numerous benchmark problems in a fraction of the runtime, and has already been applied across multiple scientific domains \citep{grinsztajn2026tabpfn3}. This makes it both a natural and highly promising candidate for predicting properties of energetic materials. The latest version (v3.5) of TabPFN can handle datasets with up to one million observations (rows) or 20,000 features (columns) \citep{jaeger2026tabpfn35}.

In this paper, we use TabPFN to predict properties of energetic materials, and demonstrate that it outperforms standard benchmark and state-of-the-art models by a substantial margin. We concentrate on impact sensitivity, crystal density and decomposition temperature, but anticipate that our results generalise to other targets as well. Compared with tree-based models that require tuning, computational cost is also drastically reduced, as TabPFN requires no additional training or fine-tuning (although its weights can be fine-tuned if desired). We apply TabPFN both directly to pre-specified sets of chemical descriptors and in combination with latent representations of molecular structures obtained from Chemprop \citep{yang2019analyzing, heid2024chemprop, graff2026chemprop}. We also compare TabPFN's built-in uncertainty estimates with the empirical distribution of the target variable for compounds with repeated observations. Finally, we demonstrate how to explain the predictions of TabPFN using Shapley values, a model-agnostic tool from XAI that ranks chemical descriptors by their contribution to the prediction.

\section{Predicting energetic material properties}
Let $y$ be a property of an energetic material, referred to as the \emph{target}, like its impact sensitivity, crystal density or decomposition temperature. When measuring $y$ requires hazardous or costly experiments, it is beneficial to predict it from a vector of chemical descriptors $x$, of length $p$, say. Unlike the target $y$, the values of $x$ are assumed to be computable, or known experimental settings, rather than measured outcomes. Thus, any machine-learning model which accurately predicts $y$ from $x$ is capable of estimating the target without any further experiments.

There are two main approaches to representing the molecule as the input vector $x$ in the energetic materials literature. Some authors, such as \citet{fournier2023in, marrs2023chemical, wu2023machine, davis2024machine, pandey2024predicting, langer2025prediction, altayeb2026new, xu2026deep}, use manual feature selection with a pre-specified list of chemical features as the predictive model's input. Others, like \citet{nguyen2021predicting, lansford2022building, hutchings2026predicting}, use tools like Chemprop's message-passing neural network to embed the molecular structure in a high-dimensional space, and use this learned embedding as the input instead. The molecular structure is originally encoded as a simplified molecular input line entry system (SMILES) string \citep{weininger1988smiles}. \citet{yang2021density, wespiser2023application} use both approaches. We use the symbol $x$ to denote the predictive model's input regardless of which of these two approaches has been employed.

\subsection{Traditional machine learning}\label{sec:trad-machine-learning}
In the traditional machine-learning framework, to predict $y$ from $x$, we start with a dataset $(\mathbf{X}, \mathbf{y})$ consisting of $N$ observations. Here, each row of the matrix $\mathbf{X}$ is a particular input $x_i$, and the entry $y_i$ of the vector $\mathbf{y}$ is the corresponding value of the target. We split the data into a training set $(\mathbf{X}_{\text{train}}, \mathbf{y}_{\text{train}})$ and a test set $(\mathbf{X}_{\text{test}}, \mathbf{y}_{\text{test}})$ of sizes $N_{\text{train}}$ and $N_{\text{test}}$, respectively, where no compound appears in both sets. The predictive model $f$ is trained on the training set (meaning, the parameters of the model are tuned so as to minimise the distance between its predictions and the values in $\mathbf{y}_{\text{train}}$) and evaluated using the test set. Its performance on the latter evaluates how well the model generalises from the training set to the test set, i.e.~how much it has learned from the training.

In this study, we chose five models to represent the most widely used and best-performing approaches in the preceding literature: the lasso, random forests, XGBoost, feed-forward neural networks and Gaussian process regression. We give a brief description of these here and refer to appropriate references for further reading. Additionally, for impact sensitivity, we consider the Variable Selection Using Random Forests (VSURF) method of~\citet{genuer2010variable}, and for decomposition temperature and crystal density, we consider the quantitative Read-Across Structure-Property Relationship (q-RASPR) models developed by~\citet{pandey2024predicting}, who also curated the versions of the two datasets we use for these tasks. As far as we are aware, these are the best-performing published models for the datasets considered in this paper. 

The least absolute shrinkage and selection operator~\citep[lasso;][]{tibshirani1996regression} uses a linear predictive model, $f(x) = \beta_0 + \beta^\top x$, where $\beta = (\beta_1, \ldots, \beta_p)^\top \in \mathbb{R}^p$. The coefficients solve the minimisation problem
\begin{equation*}
    \min_{\beta_0, \beta}\left\{\sum_{i=1}^{N_{\text{train}}}(y_i - f(x_i))^2 + \lambda\sum_{j=1}^p|\beta_j|\right\},
\end{equation*}
where $\lambda > 0$ is a tuning parameter, chosen via cross-validation on the training data. Although the lasso is not very flexible, as it is linear in the features, it is fast to compute and easy to interpret, and the regularisation shrinks the coefficients of unimportant features exactly to zero. However, when the features are highly correlated, this feature selection procedure can behave somewhat arbitrarily \citep{zou2005regularization}. Before using the lasso, we standardised the features to have mean zero and variance one, and used treatment coding with one reference level dropped for categorical features. The lasso is one of the models considered by \citet{marrs2023chemical} for predicting impact sensitivity, whose dataset we use for our model comparison in the present paper. Also, \citet{davis2024machine} found the elastic net (of which the lasso is a special case) to yield the best performance in predicting the solid-phase heat of formation of energetic materials. Following~\citet{marrs2023chemical}, we implemented the lasso using the R package \texttt{glmnet} \citep{friedman2010regularization}, version~5.1, and used default settings throughout.

The next two models, random forests \citep{breiman2001random} and XGBoost \citep{chen2016xgboost}, are ensembles of decision trees, and both produce highly flexible piecewise constant predictive models. Random forests average the predictions of many independently grown trees, each fitted to a bootstrap sample of the training data with a random subset of features at each split. XGBoost fits trees sequentially, with the next tree correcting the errors of the current ensemble. Neither model requires standardisation of features. For categorical features, we used R factors and one-hot encoding for random forests and XGBoost, respectively. VSURF, which is based on random forests, was found to be the best predictor of impact sensitivity by \citet{marrs2023chemical}, whilst \citet{davis2024machine} found XGBoost to give the most accurate predictions for crystal density. \citet{wu2023machine} found gradient boosting, of which XGBoost is a special case, to give the best predictions of decomposition temperature.

Like~\citet{marrs2023chemical}, we implemented random forests using the R package~\texttt{randomForest}~\citep{liaw2002classification}, version~4.7-1.2, with default settings throughout. For XGBoost, we used the Python library \texttt{xgboost}~\citep{chen2016xgboost}, version~3.4.2. For each nine-fold training set, we tuned the hyperparameters by comparing 30 candidate configurations in an inner 10-fold cross-validation based on the mean squared error (using~\texttt{RandomizedSearchCV} in~\texttt{scikit-learn}~\citep[version~1.9.1]{pedregosa2011scikit}). These configurations were sampled independently from a probability distribution, whose complete description is given in Table~\ref{table:xgboost-parameters} in the Appendix. The best configuration was then refitted to the entire training set and used for prediction.

Feed-forward neural networks have a long history in energetic materials prediction, going back to \citet{nefati1996prediction}, who used them to predict impact sensitivity. These networks consist of multiple layers, each of which applies an affine map (i.e.~a linear map plus a constant term) followed by an element-wise nonlinear activation function. The parameters of these affine maps, called the weights and biases of the network, are updated by gradient descent during training. As far as we are aware, no recent study has found feed-forward neural networks to outperform tree-based models when applied directly to manually selected features. However, several authors have reported that they can outperform various tree-based models when combined with message-passing neural network embeddings, such as those provided by Chemprop~\citep{yang2019analyzing, heid2024chemprop, graff2026chemprop}. When combined with a message-passing neural network, feed-forward neural networks have been shown to outperform random forests for predicting crystal density~\citep{nguyen2021predicting} and impact sensitivity~\citep{hutchings2026predicting}, as well as gradient-boosted trees for impact sensitivity~\citep{lansford2022building}. For this reason, we only include feed-forward neural networks as a contender when combined with Chemprop's embedding. Throughout, we used Chemprop's default settings, but with the inner 90/10 split for training/validation made by compound rather than by observation.

Finally, Gaussian process regression \citep{rasmussen2006gaussian} takes a Bayesian approach, placing a Gaussian process prior on the regression function, $f \sim \mathcal{GP}(\mu, k)$, with constant mean $\mu$ and covariance function $k(x, x')$ encoding assumptions about the smoothness of $f$. By definition, this means that for any finite collection of inputs $x_1, \ldots, x_m$, the vector $\mathbf{f} = (f(x_1), \ldots, f(x_m))^\top$ has a Gaussian distribution, $\mathbf{f} \sim \mathrm{N}(\mu\textbf{1}, K)$, where $\textbf{1}$ is an $m$-dimensional vector of ones and $K_{ij} = k(x_i, x_j)$. The observation model is $y_i = f(x_i) + \varepsilon_i$, where $\varepsilon_i \sim \mathrm{N}(0, \sigma^2)$, and the posterior predictive distribution at a new input is then Gaussian and available in closed form, and the posterior mean is used for prediction. We used the Mat\'{e}rn 5/2 covariance function, a common choice in practice, given by
\begin{equation*}
    k(x, x') = \gamma^2\left(1 + \frac{\sqrt{5}r}{\ell}
    + \frac{5r^2}{3\ell^2}\right)\exp\left(-\frac{\sqrt{5}r}{\ell}\right),
\end{equation*}
where $r = \lVert x - x'\rVert = \left\{\sum_{j=1}^p (x_j - x'_j)^2\right\}^{1/2}$. We used a single, isotropic length scale $\ell$, since the number of features in the datasets considered in this paper is relatively large. The parameters $\mu$, $\gamma$, $\ell$ and $\sigma^2$ were estimated by maximising the marginal likelihood of the training data. All features and targets were standardised over the training data and categorical features were one-hot encoded. Kernel ridge regression, whose predictor coincides with the posterior mean of a Gaussian process, was found by \citet{elton2018applying} to give the most accurate predictions of several computed properties on a small dataset of energetic materials.

We implemented the Gaussian processes using \texttt{GPyTorch}~\citep[version~1.15.2]{gardner2018gpytorch} on the GPU. For impact sensitivity and decomposition temperature, the smaller datasets, we maximised the marginal likelihood using the Adam optimiser with 200 steps and a learning rate of 0.1. The crystal density dataset was too large to use the exact Gaussian process, so we employed the sparse variational approximation developed by~\citet{titsias2009variational}, with 1,000 fixed inducing points placed at $k$-means centres of the training features. The variational lower bound was maximised using the Adam optimiser with 200 steps and a learning rate of 0.05. 

\subsection{Tabular foundation models}
The past few years have seen the advent of successful foundation models for tabular data, most prominently the tabular prior-data fitted network, TabPFN, by \citet{hollmann2025accurate}. TabPFN is a deep neural network based on the transformer architecture \citep{vaswani2017attention}, which has driven the remarkable performance of large language models. Unlike traditional machine-learning approaches, TabPFN is not trained on the application-specific training data $(\mathbf{X}_{\text{train}}, \mathbf{y}_{\text{train}})$. Instead, it has been pre-trained on a large collection of purely synthetic datasets, generated from randomly constructed structural causal models designed to represent a vast variety of realistic tabular datasets. It is then fed the triple $(\mathbf{X}_{\text{train}}, \mathbf{y}_{\text{train}}, \mathbf{X}_{\text{test}})$ and makes its predictions for $\mathbf{y}_{\text{test}}$ in a single forward pass, using in-context learning. That is, the training data $(\mathbf{X}_{\text{train}}, \mathbf{y}_{\text{train}})$ are given to the network as context, rather than used to update its weights. In this paper, we consider the latest two releases of TabPFN, namely versions 3 and 3.5 \citep{grinsztajn2026tabpfn3, jaeger2026tabpfn35}. We used the Python library~\texttt{tabpfn}, version~9.1.0 (apart from the analysis in Section~\ref{sec:shapley}, where we used version~8.2.0). Unless otherwise stated, the default settings were used, with the standard ensemble of eight predictors.

Since its release, TabPFN has been demonstrated to outperform state-of-the-art models in a wide variety of regression and classification tasks for tabular data. It has already been applied in a multitude of scientific disciplines (see~\citet{grinsztajn2026tabpfn3} and the references therein). As far as we are aware, this is the first study in which TabPFN is employed to predict properties of energetic materials. With no fine-tuning or weight updates required, TabPFN makes its predictions at a fraction of the time needed to tune and fit XGBoost. Furthermore, TabPFN offers an intrinsic uncertainty quantification of its predictions, as it has been pre-trained to estimate the entire distribution of $\mathbf{y}_{\text{test}}$ (conditioned on the training data), not just a point estimate. This is particularly interesting for compounds with repeated target observations, as we can compare the distribution predicted by TabPFN to the empirical one from the data.

\section{Benchmark datasets and tasks}
In this section, we describe the prediction tasks we use to compare the performance of TabPFN against the models described in Section~\ref{sec:trad-machine-learning}. We consider three separate tasks with different targets, as we are interested in evaluating TabPFN's ability to predict a range of properties of energetic materials generally, rather than a single property.

\subsection{Impact sensitivity}
The first task is to predict impact sensitivity, by which we mean the susceptibility of a material to ignite under physical impact. We use the sensitivity data provided by~\citet{marrs2023chemical}, which comprise 1533 impact sensitivity estimates (mostly from Los Alamos National Laboratory) for a total of 516 unique energetic compounds.\footnote{There is a typo in \citet{marrs2023chemical}, where it is stated that the number of unique compounds is 526, rather than 516.} Hence, these data contain many repeated measurements. In fact, nearly a third of the data are measurements for pentaerythritol tetranitrate (PETN), and the top five compounds (PETN, HMX, RDX, nitroguanidine (NQ) and trinitrotoluene (TNT)) make up more than half of the observations.

To estimate the impact sensitivity of an energetic material, a series of binary trials (typically 25 or 30) is conducted. In each, a small sample of the material is hit with a drop weight of known mass, and the operator observes whether this impact causes an explosion or not. The heights from which the weight is dropped are decided by a sequential procedure, most typically Bruceton~\citep{dixon1948method}, Neyer~\citep{neyer1994optimality} or BAM (Bundesanstalt f\"{u}r Materialforschung und -pr\"{u}fung; see~\citetalias{UN2023manual} (\citeyear{UN2023manual, UN2025model})), and the impact energy yielding an explosion with probability 50\%, called $E_{50}$ (in joules), is estimated statistically.\footnote{Technically speaking, the BAM method does not estimate $E_{50}$, but the so-called limiting stimulus of the material. Although the use of this measure of sensitivity is strongly advised against~\citep{christensen2026stop}, we do not filter out the BAM observations in the data, as we want to obtain a direct comparison with~\citet{marrs2023chemical}.} The target in the dataset of~\citet{marrs2023chemical} is $\log E_{50}$, where $\log$ denotes the base-10 logarithm.

Each compound included in the dataset is described by a list of 152 molecular descriptors computed from its structural formula (e.g.~atomic counts, atomic connectivity, oxygen balance), as well as quantum-chemical descriptors (e.g.~heat of formation, heat of explosion, HOMO--LUMO gap). Additionally, each observation contains three experimental features: first, at which laboratory the testing was performed (Los Alamos National Laboratory, the Naval Research Laboratory or other); second, whether the explosive sample was placed on a bare anvil or an anvil covered with grit paper (grit or no grit); and third, which sequential procedure was used for the binary trials (Bruceton, Neyer or BAM). We refer to~\citet{marrs2023chemical} for a complete list of all features, as well as the original data file. In the terminology of~\citet{marrs2023chemical}, we consider only the `all data' approach, not the `means data'.

\subsection{Decomposition temperature}
The second task is to predict decomposition temperature, which measures the thermal stability of a material. For this task, we use the data provided by~\citet{pandey2024predicting}, which comprise decomposition temperatures for 565 unique energetic compounds.\footnote{There is a typo in~\citet{pandey2024predicting}, where it is stated that the number of compounds is 656, rather than 565.} This dataset is a curated version of that published by~\citet{wespiser2023application}, which compiles 671 differential scanning calorimetry (DSC) measurements from the literature. The curation by~\citet{pandey2024predicting} removed duplicates and mixtures. Like the data of~\citet{marrs2023chemical}, the measurements are compiled from different sources, adding noise to the data. Furthermore, the data of~\citet{pandey2024predicting} do not contain any experimental features, so we do not know the source of each observation. Each compound is described by its SMILES string only, and the target is its decomposition temperature, given in kelvin. 

\subsection{Crystal density}
The third and final task is to predict crystal density, i.e.~mass per unit volume of a crystalline solid (measured in \unit{\gram\per\cubic\cm}). We use a larger dataset whose values originate from the Crystallography Open Database \citep[COD;][]{grazulis2009crystallography}, calculated from crystal structures obtained via X-ray crystallography. \citet{wespiser2023application} compiled a dataset from COD comprising 13,351 entries for organic compounds, which was curated further by \citet{pandey2024predicting}, averaging duplicate entries to obtain 12,805 compounds. We excluded 13 further compounds that the Python library RDKit \citep[version~2026.03.6]{rdkit} could not process because they contain silicon or phosphorus atoms with an unusually high number of bonds. This gives us a total of 12,792 compounds. Note that unlike the data used for impact sensitivity and decomposition temperature, the majority of the compounds in the crystal density dataset are not energetic (only about 9\% contain nitro, azide or peroxide groups). We decided to keep the data this way, as we want to compare our results with those obtained by \citet{pandey2024predicting}, and to ensure we evaluate the performance of TabPFN on a broad range of data.

\subsection{Model comparison}\label{sec:model-comparison}
To compare the performance of the different models, we employ the $\RMSEbar$ measure developed by~\citet{marrs2023chemical}. This has the benefit of balancing out the effect of repeated measurements of the same compound. Let $K$ denote the number of unique compounds in the dataset $\mathbf{X}$, and for each compound $k=1, 2, \ldots, K$, let $m_k$ denote the number of times it appears in $\mathbf{X}$, so that $\sum_{k=1}^K m_k = N$, the number of observations. For $0 \leq \eta \leq 1$, \citet{marrs2023chemical} define the weighted root mean square error (RMSE)
\begin{equation}\label{eq:RMSEeta}
    \mathrm{RMSE}_{\eta} = \sqrt{\sum_{k=1}^K\left\{(1 - \eta)\frac{m_k}{N} + \eta\frac{1}{K}\right\}\left(\widehat{y}_k - \bar{y}_k\right)^2},
\end{equation}
where $\widehat{y}_k$ is the predicted value for compound $k$ and $\bar{y}_k$ is the empirical average of the target values of compound $k$ in $\mathbf{X}$. If compound $k$ has repeated values in the dataset, then these are averaged in the prediction. The extreme case $\eta=0$ weights all observations equally (so a compound with repeated observations contributes more), and the case $\eta=1$ weights all compounds equally (one contribution per compound). The final measure $\RMSEbar$ is then obtained by averaging the values of $\mathrm{RMSE}_\eta$ for 1000 equally spaced values $0 \leq\eta\leq 1$. Since nearly a third of the values in the impact sensitivity data of~\citet{marrs2023chemical} are for a single compound (PETN), a balancing procedure like that in $\RMSEbar$ is crucial. Note that for datasets with no duplicate values, like the decomposition temperature and crystal density datasets, $\RMSEbar$ coincides with the standard $\mathrm{RMSE}$. This can be seen by setting $m_k = 1$ for all $k$ and $K = N$ in~\eqref{eq:RMSEeta}.

We evaluate each model using 10-fold cross-validation the same way as \citet{marrs2023chemical}. That is, we randomly split the set of compounds into ten approximately equally sized folds (so no compound has values appearing in multiple folds), and for each fold, we train the model on the data outside the fold, and use the data in the fold as test set. The folds are the same for all models. Doing this for all folds gives a prediction $\widehat{y}_k$ for each compound $k$. We then compute $\RMSEbar$ and $R^2 = 1 - (\RMSEbar / \RMSEbar_{\text{const}})^2$, where $\RMSEbar_{\text{const}}$ is the $\RMSEbar$ for the best constant prediction. For a dataset with no repeated measurements, this coincides with the standard coefficient of determination. The entire procedure is repeated ten times, and we report the average $\RMSEbar$ and $R^2$ values, along with their standard deviations.

When comparing the median runtimes of the models (per fold), we report the elapsed (i.e.~wall-clock) time, since some of the models were run on CPUs and others on GPUs, and CPU time does not account for computation on the GPU. XGBoost and Gaussian process regression can in principle be run on either a CPU or a GPU, so for these models, we did an initial pilot run to decide which was faster. The runtimes thus indicate how long it takes in practice to run each model on the hardware it is best suited for, and reflect both the method and the hardware. The pilot decided XGBoost to be run on the CPU and the Gaussian processes to be run with~\texttt{GPyTorch} on the GPU.

\subsection{Feature engineering}\label{sec:feature-engineering}
The models were compared using both manual feature selection and using Chemprop's high-dimensional molecular embedding. To explain the process by which we generated manual features, we need to consider the differences between the three datasets we use. The impact sensitivity dataset already contains a list of manually selected descriptors (along with the three experimental features), so we included a model comparison using these features directly, in the same way as in~\citet{marrs2023chemical}. The decomposition temperature and crystal density data do not, however, contain any molecular descriptors (only the SMILES strings), so we devised a list of molecular descriptors that we could apply to these data as well. We refer to this set of descriptors as the SMILES-derived descriptors. First, we computed RDKit's 2D descriptors (217 descriptors in version~2026.03.6), which are also available in Chemprop~v2. Additionally, we included those descriptors of~\citet{marrs2023chemical} that can be computed from the structural formula, without needing any quantum-chemical calculations. There were five categories of such features: element counts, atomic connectivity counts, oxygen balance, an indicator of the most sensitive functional group, and decomposition products. Finally, we removed columns with missing values, as well as constant and duplicate columns. For completeness, we ran an additional model comparison for the impact sensitivity data using this feature generation procedure, together with the three experimental features included in that dataset.

For the molecular embedding, we combined the message-passing neural network encoder from Chemprop (we used version~2.3.1 for all experiments) with the predictive models in the following way. First, in each cross-validation fold, we trained the default Chemprop model, which maps the SMILES strings to a 300-dimensional embedding via the encoder, and then maps this embedding to its prediction using a feed-forward neural network. Chemprop uses 90\% of the compounds in the training data for gradient updates, and the remaining 10\% as a validation set to pick the best epoch during training. We then froze the encoder part, and used the resulting 300-dimensional embeddings as fixed inputs for all the predictive models. For the impact sensitivity data, the three experimental features (one-hot encoded) were appended to the embedding (including during training of the default Chemprop model).

\section{Results}
In this section, we present the results of our model comparisons. Unless otherwise stated, all experiments were run on a server with two Intel Xeon E5-2620 v4 CPUs (2.10 GHz, 16 cores in total) and NVIDIA GeForce RTX 3090 GPUs, using one GPU per run.

\subsection{Impact sensitivity}\label{sec:results-impact}
For impact sensitivity, all models were compared using (i) the features from~\citet{marrs2023chemical}, (ii) the SMILES-derived descriptors from Section~\ref{sec:feature-engineering}, and (iii) the Chemprop embedding. In all three cases, the three experimental features were also included. To get a direct comparison with~\citet{marrs2023chemical} in case (i), we also include the VSURF method of~\citet{genuer2010variable} here, their best-performing model. We used the R package~\texttt{VSURF} \citep[version~1.2.2]{genuer2015vsurf}. The results are given in Tables~\ref{table:impact-marrs}, \ref{table:impact-smiles} and \ref{table:impact-chemprop}, respectively. We see that TabPFN consistently outperforms all other models considered, and that it is one of the faster models. Both v3 and v3.5 obtain similar scores, with no version consistently outperforming the other. We also see that TabPFN performs worse when given Chemprop's embedding rather than direct features.

\begin{table}[htbp]
\centering
    \caption{Comparison of the models' performance at predicting impact sensitivity using the same molecular descriptors as~\citet{marrs2023chemical}. The values are means over ten repetitions of 10-fold cross-validation, with standard deviations across repetitions in parentheses. $\RMSEbar$ is in~\si{\log\joule}. Runtimes are median times per fold. The best values are shown in bold.}
    \label{table:impact-marrs}
    \begin{tabular}{lcccccc}
    \toprule
    Model & $\RMSEbar$ && $R^2$ && Runtime (s) \\
    \midrule
    Lasso & 0.324 & (0.004) & 0.580 & (0.010) & \phantom{00}1.4 \\
    Random forest & 0.307 & (0.004) & 0.624 & (0.011) & \phantom{00}1.4 \\
    VSURF & 0.304 & (0.004) & 0.631 & (0.010) & 227.7 \\
    XGBoost & 0.320 & (0.011) & 0.588 & (0.029) & \phantom{0}27.4 \\
    Gaussian process & 0.338 & (0.006) & 0.543 & (0.015) & \phantom{00}5.1 \\
    TabPFN~v3 & \textbf{0.279} & (0.003) & \textbf{0.688} & (0.008) & \phantom{00}2.3 \\
    TabPFN~v3.5 & 0.284 & (0.004) & 0.677 & (0.010) & \phantom{00}2.2 \\
    \bottomrule
    \end{tabular}
\end{table}

\begin{table}[htbp]
\centering
    \caption{Comparison of the models' performance at predicting impact sensitivity using the SMILES-derived descriptors from Section~\ref{sec:feature-engineering}. The values are means over ten repetitions of 10-fold cross-validation, with standard deviations across repetitions in parentheses. $\RMSEbar$ is in~\si{\log\joule}. Runtimes are median times per fold. The best values are shown in bold.}
    \label{table:impact-smiles}
    \begin{tabular}{lcccccc}
    \toprule
    Model & $\RMSEbar$ && $R^2$ && Runtime (s) \\
    \midrule
    Lasso & 0.331 & (0.004) & 0.562 & (0.009)  & \phantom{0}4.6 \\
    Random forest & 0.314 & (0.005) & 0.604 & (0.011) & \phantom{0}2.0 \\
    XGBoost & 0.320 & (0.024) & 0.588 & (0.065) & 64.6 \\
    Gaussian process & 0.369 & (0.004) & 0.456 & (0.012) & \phantom{0}5.2 \\
    TabPFN~v3 & 0.293 & (0.006) & 0.656 & (0.014) & \phantom{0}3.1 \\
    TabPFN~v3.5 & \textbf{0.289} & (0.006) & \textbf{0.665} & (0.014) & \phantom{0}3.4 \\
    \bottomrule
    \end{tabular}
\end{table}

\begin{table}[htbp]
\centering
    \caption{Comparison of the models' performance at predicting impact sensitivity with Chemprop's embedding. Here, FFNN refers to Chemprop's own standard feed-forward neural network. The values are means over ten repetitions of 10-fold cross-validation, with standard deviations across repetitions in parentheses. $\RMSEbar$ is in~\si{\log\joule}. Runtimes are median times per fold and include training of Chemprop. The best values are shown in bold.}
    \label{table:impact-chemprop}
    \begin{tabular}{lcccccc}
    \toprule
    Model & $\RMSEbar$ && $R^2$ && Runtime (s) \\
    \midrule
    Lasso & 0.364 & (0.016) & 0.467 & (0.047) & \phantom{0}63.2 \\
    Random forest & 0.317 & (0.010) & 0.598 & (0.025) & \phantom{0}57.2 \\
    XGBoost & 0.318 & (0.008) & 0.596 & (0.020) & 271.4 \\
    FFNN & 0.337 & (0.015) & 0.544 & (0.042) & \phantom{0}54.1 \\
    Gaussian process & 0.350 & (0.012) & 0.510 & (0.032) & \phantom{0}59.9 \\
    TabPFN~v3 & 0.306 & (0.010) & 0.626 & (0.025) & \phantom{0}57.4 \\
    TabPFN~v3.5 & \textbf{0.303} & (0.009) & \textbf{0.632} & (0.023) & \phantom{0}57.2 \\
    \bottomrule
    \end{tabular}
\end{table}

\citet{hutchings2026predicting} considered a very similar setup to that of~\citet{marrs2023chemical}, using Chemprop to predict impact sensitivity. The results in this section are not directly comparable with theirs, however, for three reasons. First, they extended the dataset of~\citet{marrs2023chemical} with impact sensitivity data for 109 additional compounds. Second, they used transfer learning on a second dataset with 16,180 molecules, not all of them energetic. Third, they used a different weighting of the RMSE. Without further details about the data augmentation or transfer learning process, it is not possible for us to compare TabPFN's performance with their best-performing model exactly. However, we suspect that TabPFN's superior performance will carry over to their setup, considering the improvement of TabPFN applied to the Chemprop embedding over Chemprop's own predictions, as shown in Table~\ref{table:impact-chemprop}. In general, further research into combining transfer learning with tabular foundation models is encouraged.

\subsection{Decomposition temperature}\label{sec:results-temperature}
The results of the model comparison for decomposition temperature using the SMILES-derived descriptors and Chemprop's embedding are given in Tables~\ref{table:temperature-smiles} and~\ref{table:temperature-chemprop}, respectively. We see clear superior performance from TabPFN, particularly v3.5. Much like for impact sensitivity, TabPFN performs better when given the SMILES-derived descriptors than when it is combined with Chemprop's encoder.

\begin{table}[htbp]
\caption{Comparison of the models' performance at predicting decomposition temperature using the SMILES-derived descriptors from Section~\ref{sec:feature-engineering}. The values are means over ten repetitions of 10-fold cross-validation, with standard deviations across repetitions in parentheses. $\RMSEbar$ is in~\si{\kelvin}. Runtimes are median times per fold. The best values are shown in bold.}
\label{table:temperature-smiles}
\centering
\sbox0{\begin{tabular}{lccccc}
\toprule
Model & $\RMSEbar$ && $R^2$ && Runtime (s) \\
\midrule
Lasso$^\star$ & 61.1 & (0.6) & 0.531 & (0.009) & \phantom{4}6.1 \\
Random forest & 56.2 & (0.6)  & 0.603 & (0.008) & \phantom{4}0.9 \\
XGBoost & 54.2 & (1.0)  & 0.631 & (0.014) & 49.3 \\
Gaussian process & 54.6 & (0.7)  & 0.626 & (0.009) & \phantom{4}1.6 \\
TabPFN~v3 & 49.4 & (1.0)  & 0.694 & (0.012) & \phantom{4}1.3 \\
TabPFN~v3.5 & \textbf{48.8} & (1.0) & \textbf{0.701} & (0.012) & \phantom{4}1.9 \\
\bottomrule
\end{tabular}}
\begin{minipage}{\wd0}
\usebox0\par\smallskip
\footnotesize$^\star$Two of the ten repetitions had folds in which the lasso fit diverged, and these repetitions were excluded from the calculation of performance and runtimes.
\end{minipage}
\end{table}

\begin{table}[htbp]
\caption{Comparison of the models' performance at predicting decomposition temperature with Chemprop's embedding. Here, FFNN refers to Chemprop's own standard feed-forward neural network. The values are means over ten repetitions of 10-fold cross-validation, with standard deviations across repetitions in parentheses. $\RMSEbar$ is in~\si{\kelvin}. Runtimes are median times per fold and include training of Chemprop. The best values are shown in bold.}
\label{table:temperature-chemprop}
\centering
\sbox0{\begin{tabular}{lccccc}
\toprule
Model & $\RMSEbar$ && $R^2$ && Runtime (s) \\ 
\midrule
Lasso$^\star$ & 68.1 & (1.4) & 0.416 & (0.024) & \phantom{0}34.0 \\
Random forest & 58.6 & (1.1) & 0.568 & (0.017) & \phantom{0}19.2 \\
XGBoost & 58.8 & (1.1) & 0.565 & (0.017) & 207.3 \\
FFNN & 65.1 & (1.0) & 0.467 & (0.016) & \phantom{0}17.4 \\
Gaussian process & 60.0 & (1.1) & 0.548 & (0.017) & \phantom{0}19.4 \\ 
TabPFN~v3 & 55.4 & (0.8) & 0.614 & (0.011) & \phantom{0}19.2 \\
TabPFN~v3.5 & \textbf{54.1} & (1.0) & \textbf{0.633} & (0.014) & \phantom{0}19.1 \\
\bottomrule
\end{tabular}}
\begin{minipage}{\wd0}
\usebox0\par\smallskip
\footnotesize$^\star$Three of the ten repetitions had folds in which the lasso fit diverged, and these repetitions were excluded from the calculation of performance and runtimes.
\end{minipage}
\end{table}

We also evaluated the q-RASPR model developed by~\citet{pandey2024predicting}, which, to our knowledge, is the best-performing published model for these data. We used the model structure specified in \citet[Table 3]{pandey2024predicting}, with the partial least squares coefficients and read-across descriptors re-estimated on the training data for each cross-validation fold. This favours their model, as many of our test compounds are included in the data on which they performed their feature selection. The values of the structural descriptors were provided by the authors (personal communication). Using the same cross-validation folds as in the model comparison above, the q-RASPR model achieves an $\RMSEbar$ of \SI{55.2}{\kelvin} (standard deviation \SI{0.4}{\kelvin}) and $R^2$ of 0.618 (standard deviation 0.006). Hence, from Table~\ref{table:temperature-smiles}, we see that both versions of TabPFN achieve better results when given the SMILES-derived descriptors. We also applied TabPFN with the SMILES-derived descriptors to the same train-test split as used by~\citet{pandey2024predicting}, the results of which are given in Table~\ref{table:temperature-split}. Like~\citet{pandey2024predicting}, we report the RMSE, the mean absolute error (MAE) and the predictive coefficient of determination
\begin{equation*}Q^2_{\mathrm{F1}} = 1 - \frac{\sum_{i=1}^{N_{\text{test}}}(y_i - \widehat{y}_i)^2}{\sum_{i=1}^{N_{\text{test}}}(y_i - \bar{y}_{\text{train}})^2},
\end{equation*}
where both sums are over the test set. We see that both versions of TabPFN score better on all three metrics, with v3 the best overall.

\begin{table}[htbp]
\caption{Comparison of the q-RASPR model and TabPFN (with SMILES-derived descriptors) for decomposition temperature on the same train-test split as considered by~\citet{pandey2024predicting}. The values for q-RASPR are taken from~\citet{pandey2024predicting}, whereas the values for TabPFN are means over ten random seeds, with standard deviations across seeds in parentheses. RMSE and MAE are in~\si{\kelvin}. The best values are shown in bold.}
\label{table:temperature-split}
\centering
\begin{tabular}{lcccccc}
\toprule
Model & RMSE && MAE && $Q^2_{\mathrm{F1}}$ \\
\midrule
q-RASPR & 50.68 & -- & 41.38 & -- & 0.676 & -- \\
TabPFN~v3 & \textbf{49.43} & (0.40) & \textbf{37.94} & (0.35) & \textbf{0.690} & (0.005) \\
TabPFN~v3.5 & 49.94 & (0.47) & 39.06 & (0.38) & 0.683 & (0.006) \\
\bottomrule
\end{tabular}
\end{table}

\subsection{Crystal density}
The results of the model comparison for crystal density using the SMILES-derived descriptors and Chemprop's embedding are given in Tables~\ref{table:density-smiles} and \ref{table:density-chemprop}, respectively. Once again, we see that TabPFN outperforms all other models considered, with a relatively short runtime. For TabPFN~v3.5, the best-performing model, the SMILES-derived descriptors yield a better performance than Chemprop's embedding does.

\begin{table}
\caption{Comparison of the models' performance at predicting crystal density using the SMILES-derived descriptors from Section~\ref{sec:feature-engineering}. The values are means over ten repetitions of 10-fold cross-validation, with standard deviations across repetitions in parentheses. $\RMSEbar$ is in~\si{\gram\per\cm\cubed}. Runtimes are median times per fold. The best values are shown in bold.}
\label{table:density-smiles}
\centering
\begin{tabular}{lccccc}
\toprule
Model & $\RMSEbar$ && $R^2$ && Runtime (s) \\ 
\midrule
Lasso & 0.0600 & (0.0003) & 0.9025 & (0.0010) & 100.1 \\
Random forest & 0.0422 & (0.0003) & 0.9519 & (0.0006) & \phantom{0}73.4 \\
XGBoost & 0.0361 & (0.0004) & 0.9647 & (0.0008) & 405.1 \\
Gaussian process & 0.0916 & (0.0006) & 0.7732 & (0.0029) & \phantom{0}75.6 \\
TabPFN~v3 & 0.0325 & (0.0002) & 0.9715 & (0.0003) & \phantom{0}23.3 \\
TabPFN~v3.5 & \textbf{0.0314} & (0.0002) & \textbf{0.9733} & (0.0003) & \phantom{0}31.8 \\
\bottomrule
\end{tabular}
\end{table}

\begin{table}
\caption{Comparison of the models' performance at predicting crystal density with Chemprop's embedding. Here, FFNN refers to Chemprop's own standard feed-forward neural network. The values are means over ten repetitions of 10-fold cross-validation, with standard deviations across repetitions in parentheses. $\RMSEbar$ is in~\si{\gram\per\cm\cubed}. Runtimes are median times per fold and include training of Chemprop. The best values are shown in bold.}
\label{table:density-chemprop}
\centering
\begin{tabular}{lccccc}
\toprule
Model & $\RMSEbar$ && $R^2$ && Runtime (s) \\ 
\midrule
Lasso & 0.0781 & (0.0014) & 0.8351 & (0.0060) & 110.8 \\
Random forest & 0.0358 & (0.0004) & 0.9653 & (0.0008) & 156.1 \\
XGBoost & 0.0351 & (0.0004) & 0.9667 & (0.0007) & 758.9 \\
FFNN & 0.0339 & (0.0005) & 0.9689 & (0.0008) & \phantom{0}91.4 \\
Gaussian process & 0.0590 & (0.0011) & 0.9059 & (0.0034) & 176.0 \\
TabPFN~v3 & 0.0325 & (0.0003) & 0.9715 & (0.0005) & 117.7 \\
TabPFN~v3.5 & \textbf{0.0324} & (0.0003) & \textbf{0.9715} & (0.0005) & 121.9 \\
\bottomrule
\end{tabular}
\end{table}

As in Section~\ref{sec:results-temperature}, we include a comparison with the q-RASPR model of~\citet{pandey2024predicting}, which, to our knowledge, is the best-performing model for these data. The descriptor values were computed using the Python library RDKit, which reproduced their published results. Evaluating their model on the same cross-validation folds as in the model comparison yields an $\RMSEbar$ of \SI{0.0470}{\gram\per\cm\cubed} (standard deviation \SI{0.0001}{\gram\per\cm\cubed}) and an $R^2$ of 0.9403 (standard deviation 0.0003). Tables~\ref{table:density-smiles} and~\ref{table:density-chemprop} thus show that both versions of TabPFN outperform the q-RASPR model, both when given the SMILES-derived descriptors and when combined with Chemprop's embedding. As for decomposition temperature, we compare the q-RASPR model with TabPFN given the SMILES-derived descriptors on the same train-test split as~\citet{pandey2024predicting} use. The results are given in Table~\ref{table:density-split}, and show that both version of TabPFN outperform q-RASPR on all three metrics for crystal density prediction as well.

\begin{table}[htbp]
\caption{Comparison of the q-RASPR model and TabPFN (with SMILES-derived descriptors) for crystal density on the same train-test split as considered by~\citet{pandey2024predicting}. The values for q-RASPR are taken from~\citet{pandey2024predicting}, whereas the values for TabPFN are means over ten random seeds, with standard deviations across seeds in parentheses. RMSE and MAE are in~\si{\gram\per\cubic\centi\metre}. The best values are shown in bold.}
\label{table:density-split}
\centering
\begin{tabular}{lcccccc}
\toprule
Model & RMSE && MAE && $Q^2_{\mathrm{F1}}$ \\
\midrule
q-RASPR & 0.047 & -- & 0.035 & -- & 0.939 & -- \\
TabPFN~v3 & 0.03125 & (0.00007) & 0.02392 & (0.00005) & 0.9737 & (0.0001) \\
TabPFN~v3.5 & \textbf{0.03041} & (0.00007) & \textbf{0.02347} & (0.00004) & \textbf{0.9751} & (0.0001) \\
\bottomrule
\end{tabular}
\end{table}

\subsection{Additional training of Chemprop and TabPFN}
So far, when using Chemprop's embedding as input to TabPFN, we have only considered the training scheme described in Section~\ref{sec:feature-engineering}. We also conducted some additional experiments to see whether further training of Chemprop's encoder or TabPFN would lead to any improvements in accuracy. For this, we used the differentiable code path from the TabPFN library.

Let us refer to the training procedure described in Section~\ref{sec:feature-engineering} as standalone Chemprop training. Recall that this means updating the weights of both the encoder and Chemprop's default feed-forward neural network predictor, before combining the encoder with TabPFN afterwards. Leaving out standalone Chemprop training would then mean simply combining a randomly initialised Chemprop encoder with TabPFN. Once the Chemprop encoder is combined with TabPFN to form a single model, it is possible to further train the Chemprop encoder, TabPFN, or both, jointly. Here, training refers to further weight updates. By either including or excluding standalone Chemprop training, training of Chemprop's encoder and training of TabPFN, we obtain $2^3 = 8$ configurations to compare.

We evaluated the eight different configurations on the impact sensitivity dataset of~\citet{marrs2023chemical} via a 10-fold cross-validation, similar to that described in Section~\ref{sec:model-comparison}. However, for each test fold, we split the compounds in the nine remaining folds into a training set (90\% of compounds) and a validation set (10\% of compounds), the latter of which was used for early stopping and to select the best epoch. When the model was evaluated on the test fold, the validation compounds were added back to the training compounds as context for TabPFN.

Unlike standard machine-learning models, TabPFN takes entire datasets as input. Therefore, we performed only one training step per epoch, using the entire training set each time. In each step, we randomly selected 80\% of the training set observations to provide as context with both features and labels, and updated the weights to minimise the mean squared error of TabPFN's predictions for the remaining 20\%. Early stopping was based on validation mean squared error with a patience of ten epochs, and a maximum of 50. For fine-tuning, we used the AdamW optimiser~\citep{loshchilov2018decoupled} with a learning rate of $10^{-4}$ for Chemprop weights and $10^{-6}$ for TabPFN weights. We used a weight decay of $10^{-5}$ and clipped gradients to at most norm 1. For numerical stability, we skipped all training steps where the loss or gradient norm was not a finite number, which can happen when TabPFN~v3.5 receives columns with constant values. For standalone Chemprop training, we used default parameters, 50 epochs, 64 observations per batch and norm aggregation, keeping the weights from the epoch with the best validation loss. 

To save time, we used a single TabPFN predictor rather than the default choice of an ensemble of eight used in the previous experiments. In addition, before training, the measurements in the dataset were averaged within all combinations of the tuple (compound, laboratory, grit, method), so that each unique combination contributed a single training observation. $\RMSEbar$ was nevertheless computed per compound, as in Section~\ref{sec:model-comparison}. For these reasons, the results of this section are not directly comparable with those of Table~\ref{table:impact-chemprop}.

\begin{table}[htbp]
    \centering
    \caption{Comparison of different training configurations of Chemprop's encoder and differentiable TabPFN (v3.5). Values are means over ten repetitions of 10-fold cross-validation, with standard deviations across repetitions in parentheses. The final column shows the number of repetitions in which the configuration achieved a lower $\RMSEbar$ than the no-training configuration within the same block. $\RMSEbar$ is in~\si{\log\joule}. Runtimes are median times per fold. The runtimes in the first block include the time spent on standalone Chemprop training. The best values are shown in bold.}
    \label{table:results_finetuning_v3.5}
    \begin{tabular}{ccccccc}
        \toprule
        \shortstack{Standalone\\Chemprop training} & \shortstack{Trained\\Chemprop} & \shortstack{Trained\\TabPFN} & $\RMSEbar$ & $R^2$ & Runtime (s) & \shortstack{Better than\\no training} \\
        \midrule
        $\checkmark$ &&& \cellstack{0.301}{(0.009)} & \cellstack{0.636}{(0.022)} & 23.0 & -- \\
        $\checkmark$ && $\checkmark$ & \cellstack{0.302}{(0.010)} & \cellstack{0.635}{(0.024)} & 37.3 & \phantom{1}5/10 \\
        $\checkmark$ & $\checkmark$ && \cellstack{0.301}{(0.012)} & \cellstack{0.637}{(0.029)} & 28.7 & \phantom{1}4/10 \\
        $\checkmark$ & $\checkmark$ & $\checkmark$ & \cellstack{0.301}{(0.010)} & \cellstack{0.637}{(0.024)} & 29.8 & \phantom{1}4/10 \\
        \midrule
        &&& \cellstack{0.296}{(0.007)} & \cellstack{0.649}{(0.018)} & \phantom{0}1.4 & -- \\
        && $\checkmark$ & \cellstack{0.295}{(0.007)} & \cellstack{0.652}{(0.018)} & 13.8 & \phantom{1}9/10 \\
        & $\checkmark$ && \cellstack{\textbf{0.294}}{(0.007)} & \cellstack{\textbf{0.653}}{(0.017)} & \phantom{0}5.9 & 10/10 \\
        & $\checkmark$ & $\checkmark$ & \cellstack{\textbf{0.294}}{(0.007)} & \cellstack{\textbf{0.653}}{(0.017)} & \phantom{0}7.1 & 10/10 \\
        \bottomrule
    \end{tabular}
\end{table}

The results are given in Table~\ref{table:results_finetuning_v3.5}. Overall, the differences in $\RMSEbar$ are relatively small, indeed smaller than the standard deviations. When standalone Chemprop training is left out, we do, however, see that further training of Chemprop and/or TabPFN leads to an improvement. This is clear from examining the paired comparisons per repetition in the right-most column. Note also that runtime is substantially reduced when leaving out standalone Chemprop training.

Rather surprisingly, standalone Chemprop training seems to have a slightly negative impact on performance, regardless of the other training configurations. Even with a completely untrained Chemprop encoder, TabPFN achieves comparable and even better results, which is impressive. This suggests that the encoder's message-passing neural network architecture, even with random weights, is capturing some information about the molecules' structure that TabPFN is able to use. As for the negative impact of standalone training, one possible explanation is that when Chemprop's encoder is trained together with its feed-forward neural network predictor, it somehow discards information that otherwise TabPFN could have exploited. When this counter-intuitive result revealed itself, we suspected that it could be an effect of Chemprop not having enough data for proper standalone training. We therefore evaluated the effect of standalone Chemprop training on all three datasets. To ensure a direct comparison, we used the same default TabPFN settings and cross-validation folds as in Tables~\ref{table:impact-chemprop}, \ref{table:temperature-chemprop} and~\ref{table:density-chemprop}. The results are given in Table~\ref{table:untrained-chemprop}, and support our hypothesis. Although standalone training has little effect on the results for impact sensitivity with TabPFN~v3, leaving out training yields better results in eight out of ten repetitions for version 3.5. For decomposition temperature, the pattern is more striking: the untrained encoder achieves a lower $\RMSEbar$ than the trained encoder in all ten repetitions. For crystal density, where the dataset is much larger, the opposite is true: standalone Chemprop training improves the performance in all ten repetitions.

\begin{table}[htbp]
    \centering
    \caption{Comparison of standalone Chemprop training versus no training, for the three datasets. The values are means over ten repetitions of 10-fold cross-validation, with standard deviations across repetitions in parentheses. $\RMSEbar$ is in~\si{\log\joule} for impact sensitivity, \si{\kelvin} for decomposition temperature and \si{\gram\per\cm\cubed} for crystal density. Runtimes are median times per fold, including standalone Chemprop training where applicable. The final column gives the number of repetitions in which the untrained encoder has the lower $\RMSEbar$.}
    \label{table:untrained-chemprop}
    \begin{tabular}{ccccccc}
    \toprule
    Dataset & \cellstack{TabPFN}{version} & \cellstack{Chemprop}{encoder} & $\RMSEbar$ & $R^2$ & Runtime (s) & \cellstack{Untrained}{better} \\
    \midrule
    \multirow{6}{*}{\cellstack{Impact}{sensitivity}}
        & 3\phantom{.5} & Trained   & \cellstack{0.306}{(0.010)} & \cellstack{0.626}{(0.025)} & 57.4 & \\
        & 3\phantom{.5} & Untrained & \cellstack{0.306}{(0.010)} & \cellstack{0.625}{(0.025)} & \phantom{0}3.0 & \phantom{1}5/10 \\
        & 3.5           & Trained   & \cellstack{0.303}{(0.009)} & \cellstack{0.632}{(0.023)} & 57.2 & \\
        & 3.5           & Untrained & \cellstack{0.295}{(0.008)} & \cellstack{0.651}{(0.020)} & \phantom{0}2.9 & \phantom{1}8/10 \\
    \midrule
    \multirow{6}{*}{\cellstack{Decomposition}{temperature}}
        & 3\phantom{.5} & Trained   & \cellstack{55.4}{(0.8)} & \cellstack{0.614}{(0.011)} & 19.2 & \\
        & 3\phantom{.5} & Untrained & \cellstack{53.6}{(1.1)} & \cellstack{0.639}{(0.015)} & \phantom{0}1.1 & 10/10 \\
        & 3.5           & Trained   & \cellstack{54.1}{(1.0)} & \cellstack{0.633}{(0.014)} & 19.1 & \\
        & 3.5           & Untrained & \cellstack{52.8}{(1.1)} & \cellstack{0.650}{(0.015)} & \phantom{0}1.2 & 10/10 \\
    \midrule
    \multirow{6}{*}{\cellstack{Crystal}{density}}
        & 3\phantom{.5} & Trained   & \cellstack{0.0325}{(0.0003)} & \cellstack{0.9715}{(0.0005)} & 117.7 & \\
        & 3\phantom{.5} & Untrained & \cellstack{0.0383}{(0.0007)} & \cellstack{0.9602}{(0.0014)} & \phantom{0}13.6 & \phantom{1}0/10 \\
        & 3.5           & Trained   & \cellstack{0.0324}{(0.0003)} & \cellstack{0.9715}{(0.0005)} & 121.9 & \\
        & 3.5           & Untrained & \cellstack{0.0346}{(0.0003)} & \cellstack{0.9676}{(0.0005)} & \phantom{0}18.3 & \phantom{1}0/10 \\
    \bottomrule
    \end{tabular}
\end{table}

\section{Analysis and Interpretation}
Having demonstrated the superior performance of TabPFN in our model comparisons, we briefly consider how its predictions can be analysed and explored further. This is not the main focus of the present paper, and we do not make any attempt at an exhaustive account. The purpose of this section is rather to present a proof of concept, and hopefully to inspire future research questions. We restrict ourselves to considering impact sensitivity, focusing on TabPFN~v3 when given the features from~\citet{marrs2023chemical}, the best-performing model on these data.

\subsection{Feature importance with Shapley values}\label{sec:shapley}
The predictions of TabPFN~v3 are straightforward to explain using the companion package
\texttt{tabpfn-extensions}\footnote{\url{https://github.com/PriorLabs/tabpfn-extensions}}, which provides several XAI tools for this purpose. We use Shapley values \citep{shapley1953value} from cooperative game theory. These split the difference between the prediction and a baseline into contributions from the individual features, so we can see which features drive the prediction and in which direction. The baseline is the model's average predicted $\log E_{50}$ over the context. We assume the features to be independent, which is clearly unrealistic but a pragmatic choice given the computational complexity associated with the high number of features in the data. An interesting avenue for future research is to incorporate the dependence between the features. This will probably need to involve some feature reduction procedure to be computationally feasible. The experiments of this section were computed on an NVIDIA RTX 4000 Ada Generation GPU. Full details about how the Shapley values were implemented can be found in Section~\ref{app:shapley_computation} in the Appendix.

We illustrate Shapley values on PETN, the compound with the most measurements in the impact sensitivity data (490 in total). To mimic the prediction of a new material, we give TabPFN the observations of all other compounds as context, so PETN itself is not part of the training data. All 490 PETN observations have the same values of the chemical descriptors and differ only in the three experimental features. We compute Shapley values for every observation and average them over the 490 observations. The same analysis for the next four most numerous compounds, namely HMX, RDX, NQ and TNT, can be found in Section~\ref{app:shapley_molecules} in the Appendix. Although they performed a global feature attribution analysis, it is interesting to compare our results with those of~\citet{marrs2023chemical}, since some of the same patterns emerge. This is especially true for PETN and TNT, where the four most important features found by~\citet{marrs2023chemical} appear among our top five features by Shapley value.

\begin{figure}[htbp]
    \centering
    \begin{subfigure}{0.495\textwidth}
        \includegraphics[width=\linewidth]{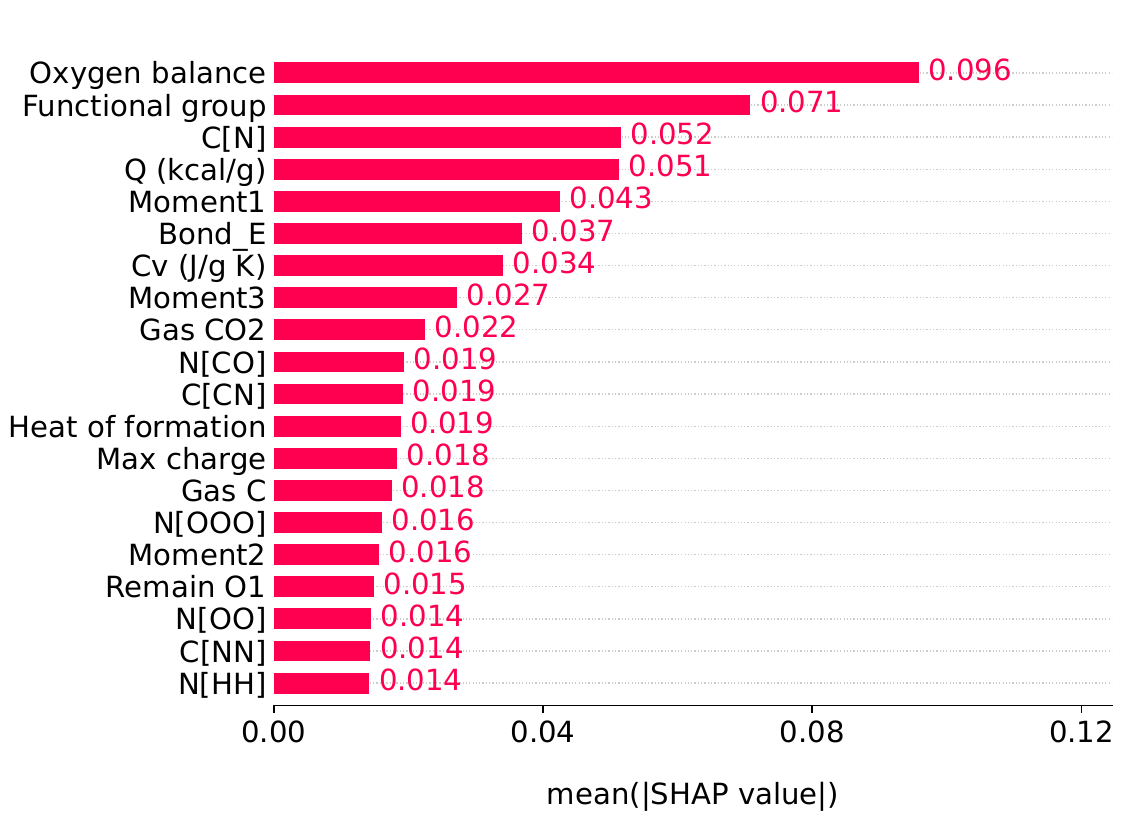}
    \end{subfigure}
    \hfill
    \begin{subfigure}{0.495\textwidth}
        \includegraphics[width=\linewidth]{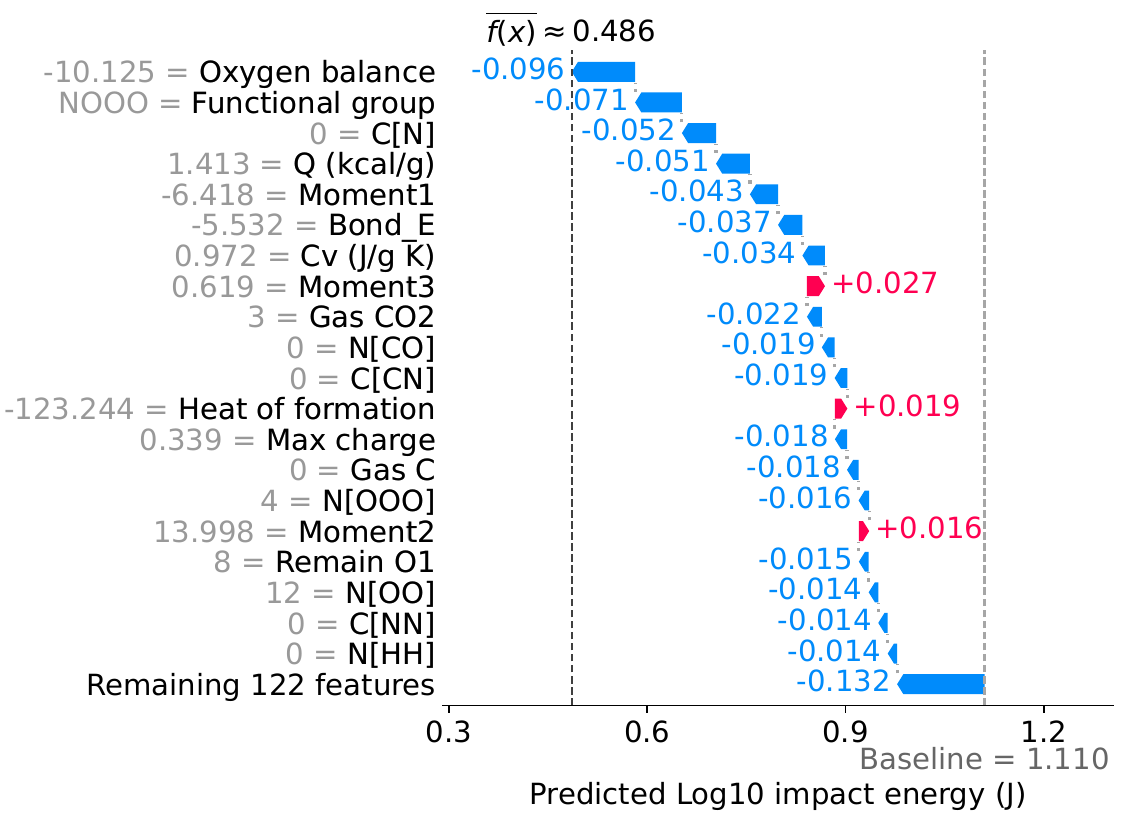}
    \end{subfigure}
    \caption{Shapley values for PETN, averaged over its 490 observations, for the twenty
    chemical descriptors with the largest mean absolute Shapley values. The bottom row
    groups the remaining 122 features, including the three experimental features.}
    \label{fig:petn_shap}
\end{figure}

Figure~\ref{fig:petn_shap} presents the twenty chemical descriptors with the largest mean absolute Shapley values. The left panel measures how important each descriptor is on average, while the right panel averages the signed Shapley values instead and plots them as a waterfall. Starting from the baseline at the bottom, the arrows add the contributions one by one and end at the average prediction for PETN at the top, so the waterfall summarises the compound rather than explaining a single observation. The grey labels give PETN's descriptor values. As the target is $\log E_{50}$, a positive (red) contribution increases the predicted impact energy, meaning lower predicted sensitivity, whereas a negative (blue) contribution means higher predicted sensitivity. The bottom row groups the remaining 122 features, including the three experimental features.

The average prediction for PETN, $\log E_{50} = 0.486$, lies well below the baseline of 1.110, so the model predicts PETN to be more sensitive than the average observation in its context. Oxygen balance and functional group are the most important descriptors, and both push the prediction towards higher sensitivity. Each of the descriptors contributes in the same direction in all 490 observations, which is why the bars in the left panel match the arrows in the right panel.
Note that these attributions describe the behaviour of the model, not causal chemical effects.

\subsection{The full predictive distribution}
A feature of TabPFN that sets it apart from traditional machine-learning methods is that it estimates the entire predictive distribution of its target, rather than simply a point estimate. Since the impact sensitivity dataset by~\citet{marrs2023chemical} contains repeated measurements for several compounds, it is possible for us to compare these predictive distribution estimates against the corresponding empirical ones from the data. We do so for the five most numerous compounds, given in Table~\ref{table:compound-counts}.

\begin{table}[htbp]
    \centering
    \caption{The number of observations, unique combinations of experimental features and
    context observations for the five most numerous compounds in the data of~\citet{marrs2023chemical}.}
    \label{table:compound-counts}
    \begin{tabular}{lrrrrr}
        \toprule
         & PETN & HMX & RDX & NQ & TNT \\
        \midrule
        Observations & 490 & 164 & 130 & 44 & 40 \\
        Unique combinations & 8 & 7 & 9 & 3 & 6 \\
        Context observations & 1043 & 1369 & 1403 & 1489 & 1493 \\
        \bottomrule
    \end{tabular}
\end{table}

\begin{figure}[htbp]
    \includegraphics[width=\linewidth]{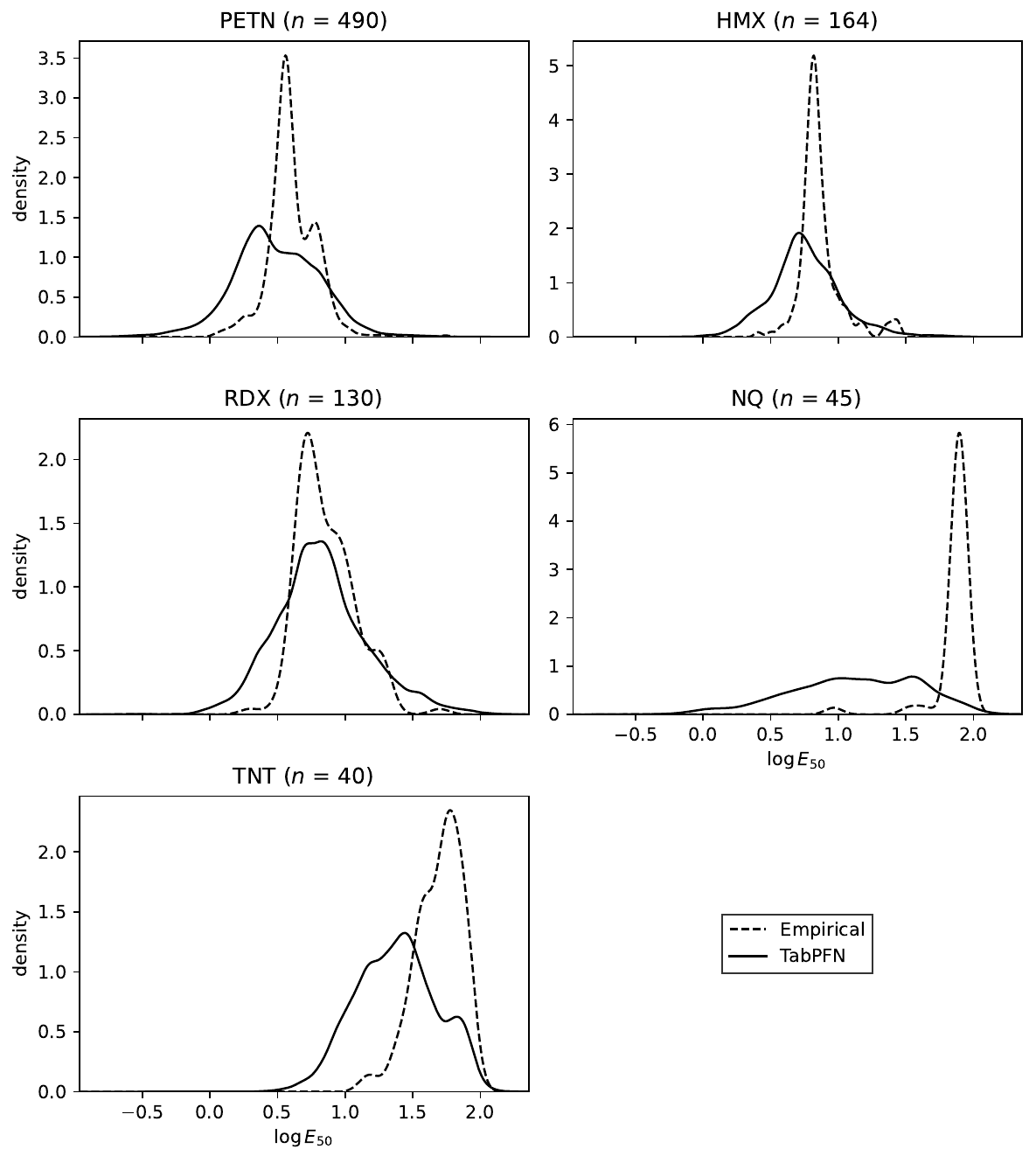}
    \caption{Kernel density estimates of the full predictive distributions of TabPFN for the five most numerous compounds, and for their empirical distributions from the data.}
    \label{fig:full-predictive}
\end{figure}

For each of the five compounds, we use the data for all other compounds as context. To get a single distribution for a given compound, we average the predictive distributions over the combinations of the three experimental features, weighted by their proportions among the compound's measurements. The kernel density estimates are generated using R's default kernel bandwidth and with inverse-transform sampling using TabPFN's quantiles. The results are given in Figure~\ref{fig:full-predictive}. The agreement is strongest for PETN, HMX and RDX, the three compounds with the most measurements in the data. TabPFN predicts both NQ and TNT to be more sensitive than their empirical data suggest.

\section{Conclusion}
We have used TabPFN, a new tabular foundation model, to predict three properties of energetic materials: impact sensitivity, decomposition temperature and crystal density. The three datasets vary in their size, and the first contains many repeated measurements for the same compounds, as well as three experimental features. Our results show that TabPFN (versions 3 and 3.5) consistently outperforms penalised linear regression (the lasso), tree-based models (random forests and XGBoost), feed-forward neural networks and Gaussian processes. This holds both when the models are given a list of chemical descriptors and when combined with Chemprop's embedding. Overall, TabPFN achieved better results when using chemical descriptors as input rather than Chemprop's embedding. 

We also compared TabPFN directly against (to our knowledge) the best published models on the three datasets, namely VSURF for impact sensitivity and q-RASPR for decomposition temperature and crystal density. When given chemical descriptors, TabPFN achieves more accurate results in all cases. 

Further training of Chemprop's encoder and/or TabPFN had little effect on performance (but did increase runtimes), according to our experiment on the impact sensitivity data. As a proof of concept, we compared TabPFN's distribution predictions with the empirical distributions of five compounds with many repeated measurements in the impact sensitivity dataset. We also computed Shapley values to explain TabPFN's predictions of impact sensitivity for these compounds.

We did not study the effect of transfer learning, as implemented by~\citet{lansford2022building} and \citet{hutchings2026predicting}, but consider combining TabPFN with transfer learning a fruitful path for future research.

\section*{Acknowledgements} The work by Lars H.\ B.\ Olsen is funded by the Research Council of Norway's Centre of Excellence Integreat (project no.\ 332645).

\section*{Use of generative AI}
The authors used the model Claude Opus~5.5 by Anthropic for assistance in literature review, coding, proofreading and checking the manuscript for mistakes and inconsistencies. The authors reviewed and edited all output and take full responsibility for the content of the paper.

\clearpage
\appendix
\section*{Appendix}

\section{Details about XGBoost}
\begin{table}[htbp]
    \centering
    \caption{The search space for the XGBoost hyperparameters. In each outer training set, 30 candidate configurations were drawn independently from the distributions indicated. $\mathrm{U}(a, b)$ denotes the uniform distribution on $[a, b]$ and $\log\mathrm{U}(a, b)$ the log-uniform distribution. Integers are sampled uniformly, with end points included. The default column gives the default value in the Python library~\texttt{xgboost}.}
    \label{table:xgboost-parameters}
    \begin{tabular}{lcl}
    \toprule
    Hyperparameter & Default & Distribution \\
    \midrule
    Number of trees                  & 100 & $\{100, \ldots, 1000\}$ \\
    Learning rate                    & 0.3 & $\log\mathrm{U}(0.01, 0.3)$ \\
    Maximum tree depth               & 6   & $\{2, \ldots, 8\}$ \\
    Minimum child weight             & 1   & $\{1, \ldots, 10\}$ \\
    Row subsampling rate             & 1   & $\mathrm{U}(0.6, 1)$ \\
    Column subsampling rate per tree & 1   & $\mathrm{U}(0.5, 1)$ \\
    $L_2$ penalty                    & 1   & $\log\mathrm{U}(10^{-3}, 10)$ \\
    $L_1$ penalty                    & 0   & $\log\mathrm{U}(10^{-4}, 1)$ \\
    \bottomrule
    \end{tabular}
\end{table}

\section{Shapley value analysis}
\label{app:shapley}

This appendix describes how we computed the Shapley values in Section~\ref{sec:shapley} and
shows the results for four more compounds.

\subsection{Computing the Shapley values}
\label{app:shapley_computation}

We compute the Shapley values with \texttt{tabpfn-extensions}, whose explainers are built on
the \texttt{shapiq} package \citep{muschalik2024shapiq}.

\paragraph{One model per compound.}
We explain the five compounds with the most observations: PETN, HMX, RDX, NQ and TNT.
For each of them, we recontextualise TabPFN~v3, that is, we
give it the observations of all other compounds as context.
The explained compound is therefore never part of the context, and each compound gets its
own baseline.
We use eight predictors and fit the model with \path{fit_mode="fit_with_cache"}. The context caching speeds up the Shapley value computations described below.

\paragraph{One explanation per combination of experimental features.}
Within a compound, all observations have the same values of the chemical features and
differ only in the three experimental features.
Observations with the same combination of these three features therefore have identical
inputs, so their predictions and Shapley values are also identical.
We explain each unique combination once and copy its Shapley values to all observations with
that combination.
This reduces the number of explanations from 868 to 33 (see Table~\ref{table:compound-counts}),
which saves computation without changing the results.
The averages in the figures are taken over observations rather than combinations, so
frequent combinations weigh more.
For PETN, for example, 226 of the 490 observations were tested at Los Alamos National
Laboratory with grit paper and the Bruceton procedure.

\paragraph{Approximate marginal Shapley values.}
Let $\mathcal{M} = \{1, \dots, M\}$ be the set of features and $v(S)$ the prediction when
only the features in a subset $S \subseteq \mathcal{M}$, called a coalition, are known.
The Shapley value of feature $j$ is
$\phi_j = \sum_{S \subseteq \mathcal{M} \setminus \{j\}} \frac{|S|!\,(M - |S| - 1)!}{M!}
\bigl(v(S \cup \{j\}) - v(S)\bigr)$.
Each term is the change in the prediction when $j$ is revealed after the features in $S$,
weighted by the fraction of orders in which exactly these features come before $j$.
Exact Shapley values are expensive to compute, especially with as many features as here,
because they require $v(S)$ for every coalition.
After removing all constant columns, we have $M = 142$ features, which means there are $2^{142} \approx 5.6 \times 10^{42}$ coalitions.
We therefore approximate the Shapley values with OddSHAP \citep{fumagalli2026odd}, the
estimator that \texttt{tabpfn-extensions} recommends for TabPFN, with a budget of 25,000 coalitions for each combination of experimental features.

There are several ways to define $v(S)$.
Conditional Shapley values average the prediction over the unknown features, drawn from
their distribution given the known ones.
This respects the dependence between the features, but requires a conditional distribution
for every coalition, which is costly to estimate
\citep{olsen2024comparative, olsen2026computing}.
To speed up the computations further, we use marginal Shapley values, which ignore this
dependence.

For marginal Shapley values, we estimate the value of a non-empty coalition $S$ as $v(S) = \frac{1}{T} \sum_{k=1}^{T} f\bigl(x_S, z^{(t)}_{\bar{S}}\bigr)$, where $f$ is the TabPFN prediction of $\log E_{50}$. In each prediction, we keep the known feature values $x_S$ fixed and fill in the unknown features in $\bar{S}$ jointly from a background observation $z^{(t)}$. We draw $T = 250$ such observations at random from the context without replacement. For the empty coalition, the \texttt{shapiq} implementation instead averages the predictions over the whole context. The baseline is therefore $v(\emptyset) = \frac{1}{N} \sum_{i=1}^{N} f\bigl(x^{(i)}\bigr)$, where $x^{(1)}, \dots, x^{(N)}$ are the $N$ context observations. 
With $25{,}000$ coalitions and $250$ background observations, coalition evaluation requires $6.25 \times 10^6$ model predictions per combination of experimental features. The context caching described above speeds up these computations.

\paragraph{Software.}
We use \texttt{tabpfn} 8.2.0, \texttt{tabpfn-extensions} 0.5.0,
\texttt{shapiq} 1.6.0 and the TabPFN checkpoint \texttt{tabpfn-\allowbreak v3-\allowbreak regressor-\allowbreak v3\_\allowbreak default.ckpt}.
All random seeds are set to 0.

\subsection{Shapley values for HMX, RDX, NQ and TNT}
\label{app:shapley_molecules}

\begin{figure}[htbp]
    \centering
    \begin{subfigure}{0.495\textwidth}
        \includegraphics[width=\linewidth]{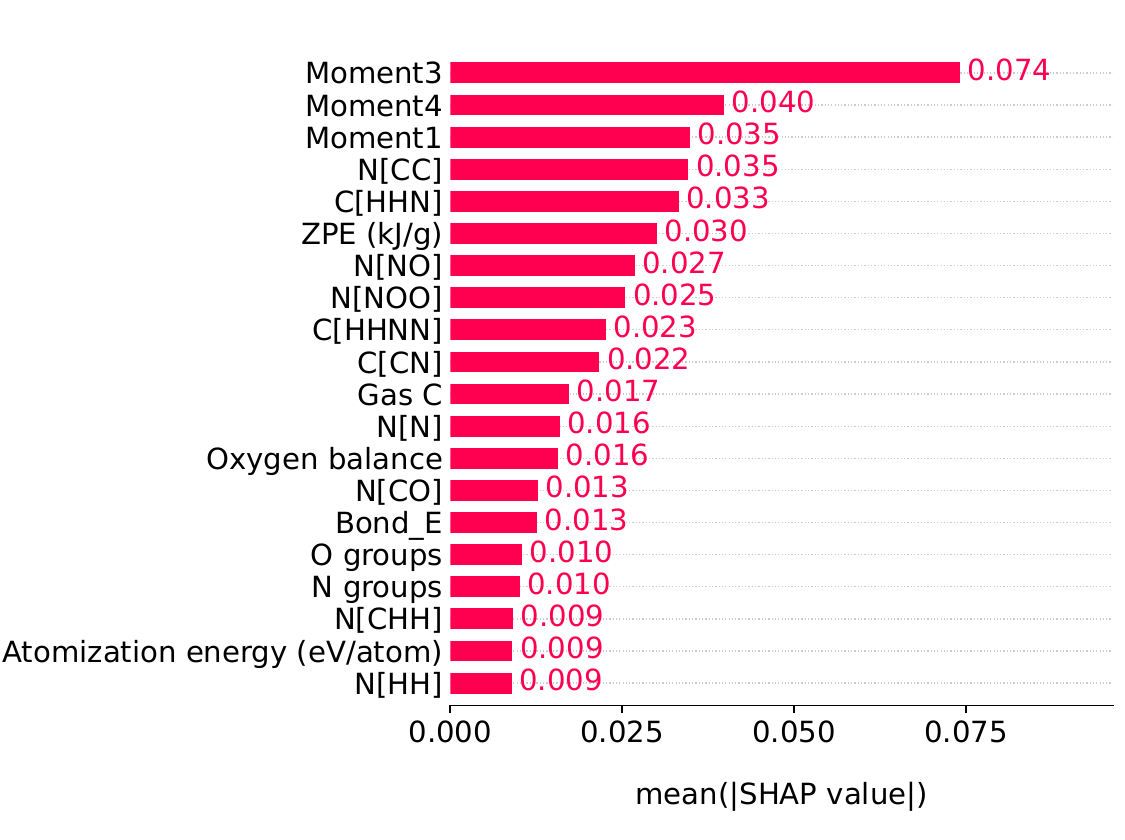}
    \end{subfigure}
    \hfill
    \begin{subfigure}{0.495\textwidth}
        \includegraphics[width=\linewidth]{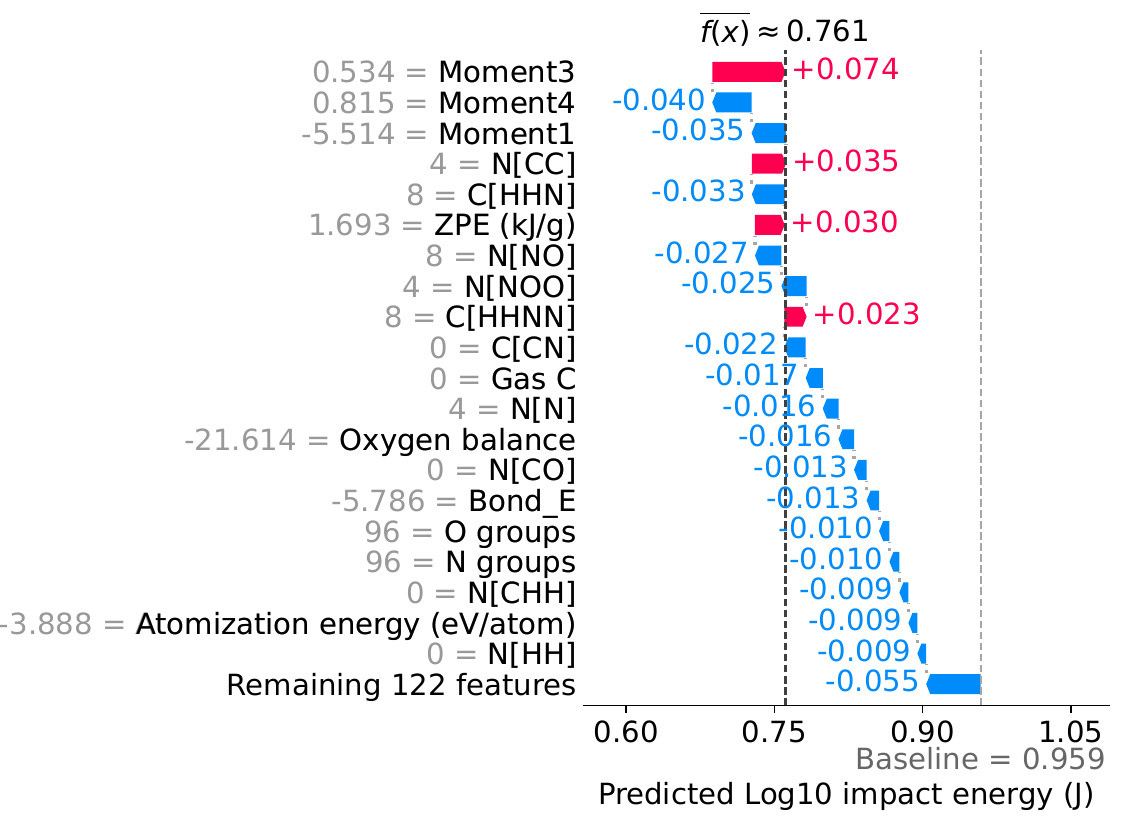}
    \end{subfigure}
    \caption{Shapley values for HMX, averaged over its 164 observations. The panels are as
    in Figure~\ref{fig:petn_shap}.}
    \label{fig:hmx_shap}
\end{figure}

\begin{figure}[htbp]
    \centering
    \begin{subfigure}{0.495\textwidth}
        \includegraphics[width=\linewidth]{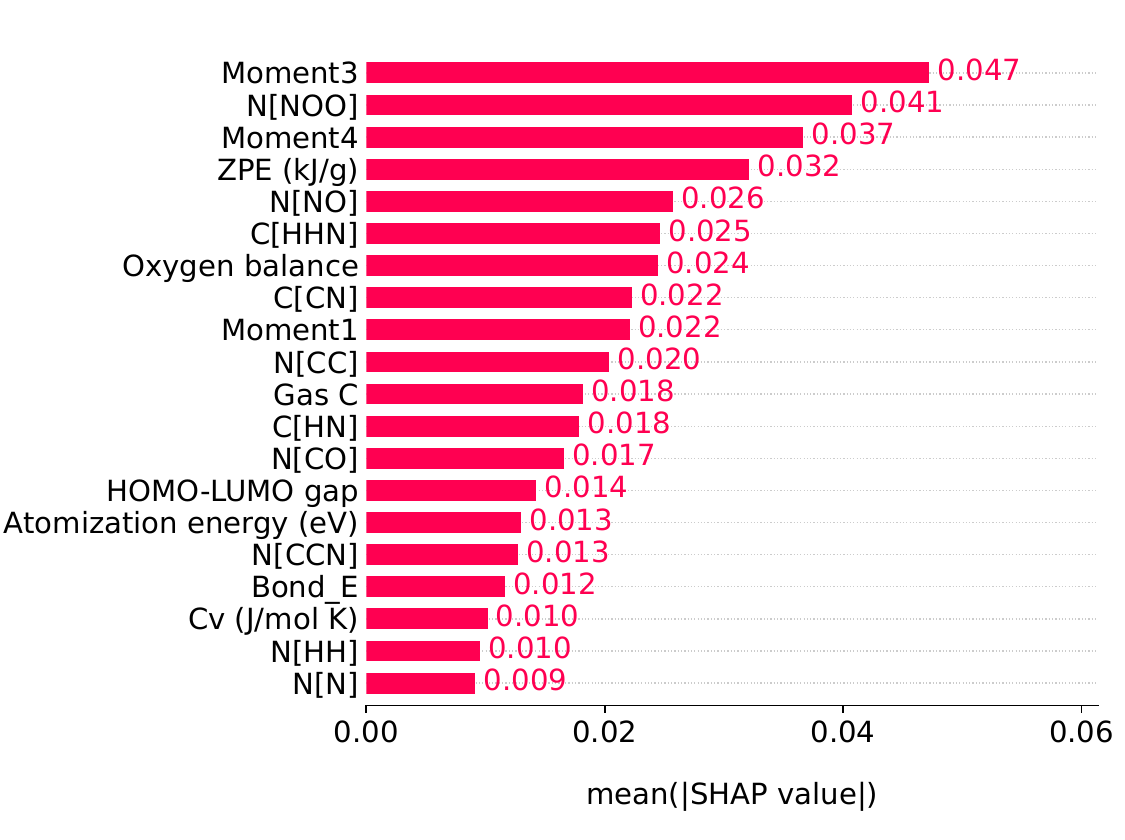}
    \end{subfigure}
    \hfill
    \begin{subfigure}{0.495\textwidth}
        \includegraphics[width=\linewidth]{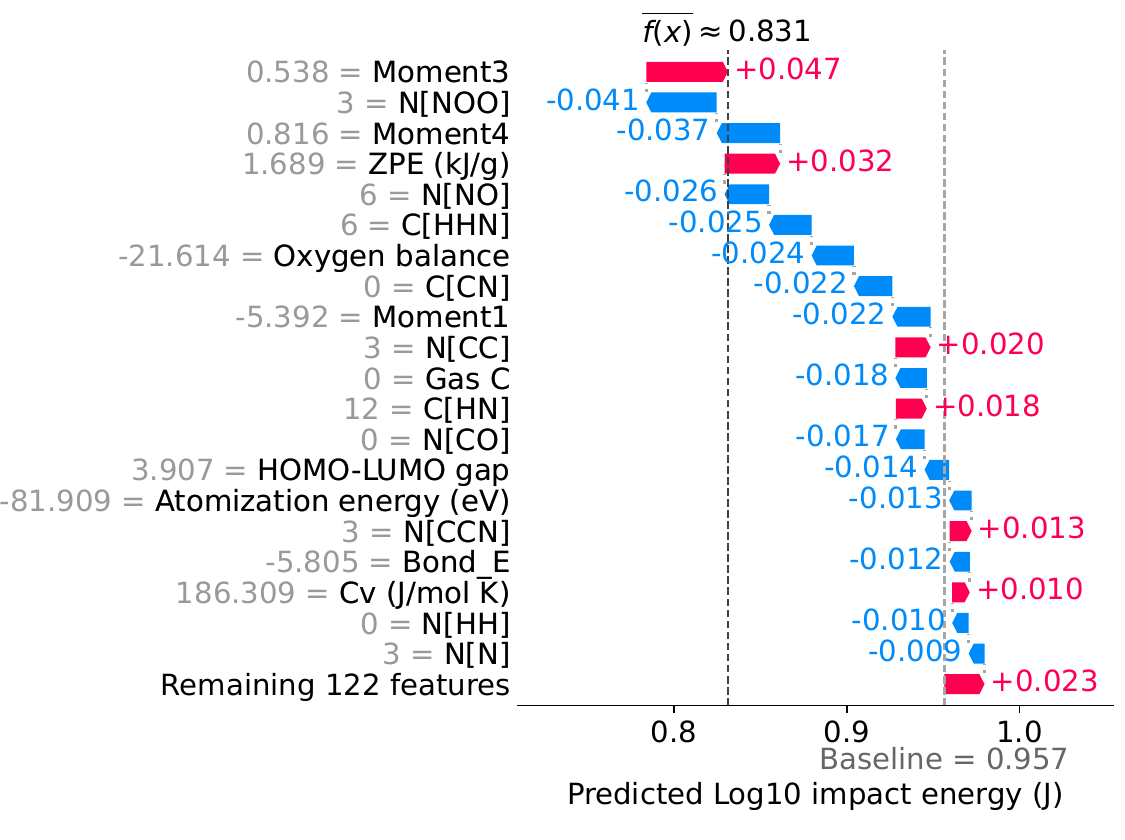}
    \end{subfigure}
    \caption{Shapley values for RDX, averaged over its 130 observations. The panels are as
    in Figure~\ref{fig:petn_shap}.}
    \label{fig:rdx_shap}
\end{figure}

\begin{figure}[htbp]
    \centering
    \begin{subfigure}{0.495\textwidth}
        \includegraphics[width=\linewidth]{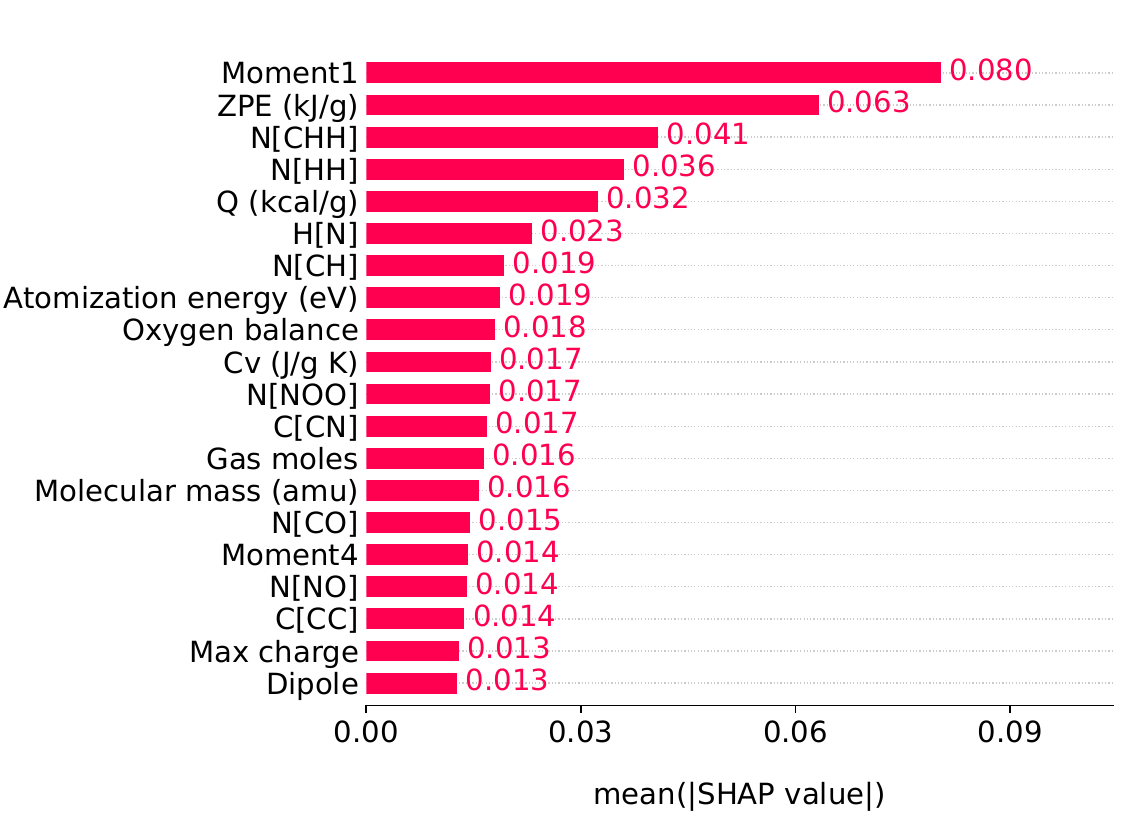}
    \end{subfigure}
    \hfill
    \begin{subfigure}{0.495\textwidth}
        \includegraphics[width=\linewidth]{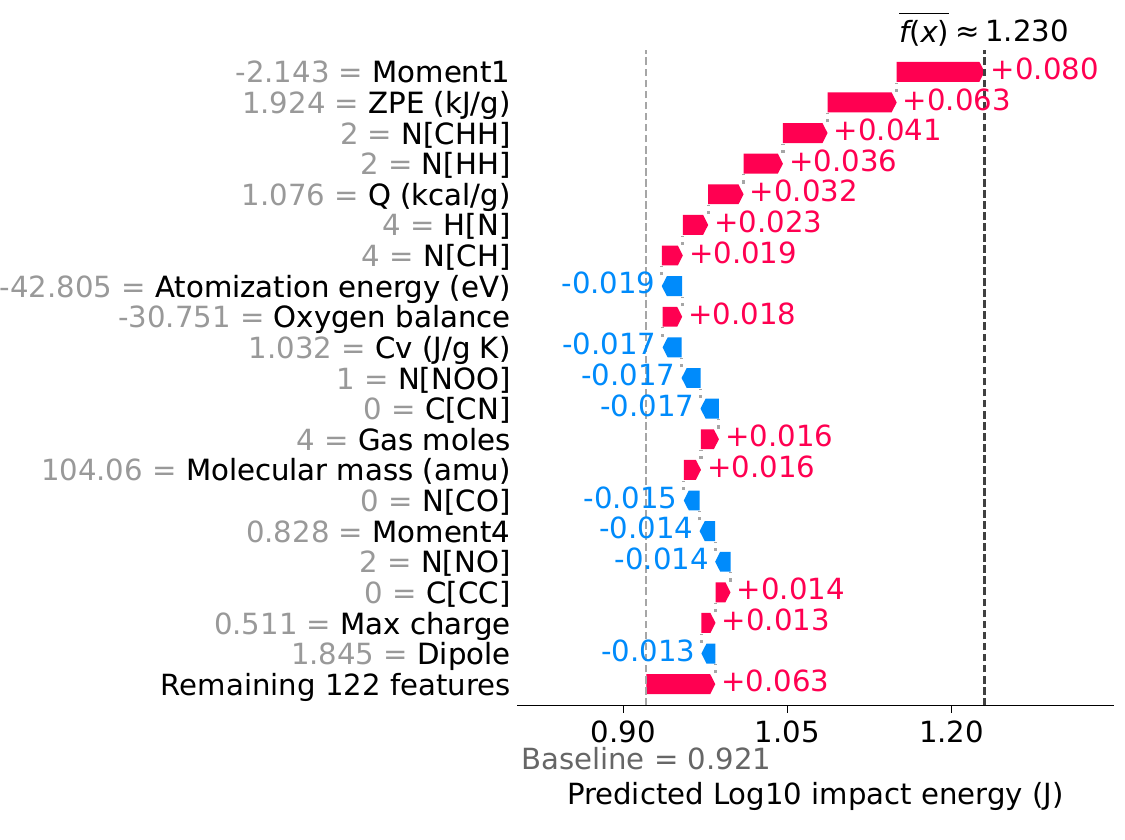}
    \end{subfigure}
    \caption{Shapley values for NQ, averaged over its 44 observations. The panels are as
    in Figure~\ref{fig:petn_shap}.}
    \label{fig:nq_shap}
\end{figure}

\begin{figure}[htbp]
    \centering
    \begin{subfigure}{0.495\textwidth}
        \includegraphics[width=\linewidth]{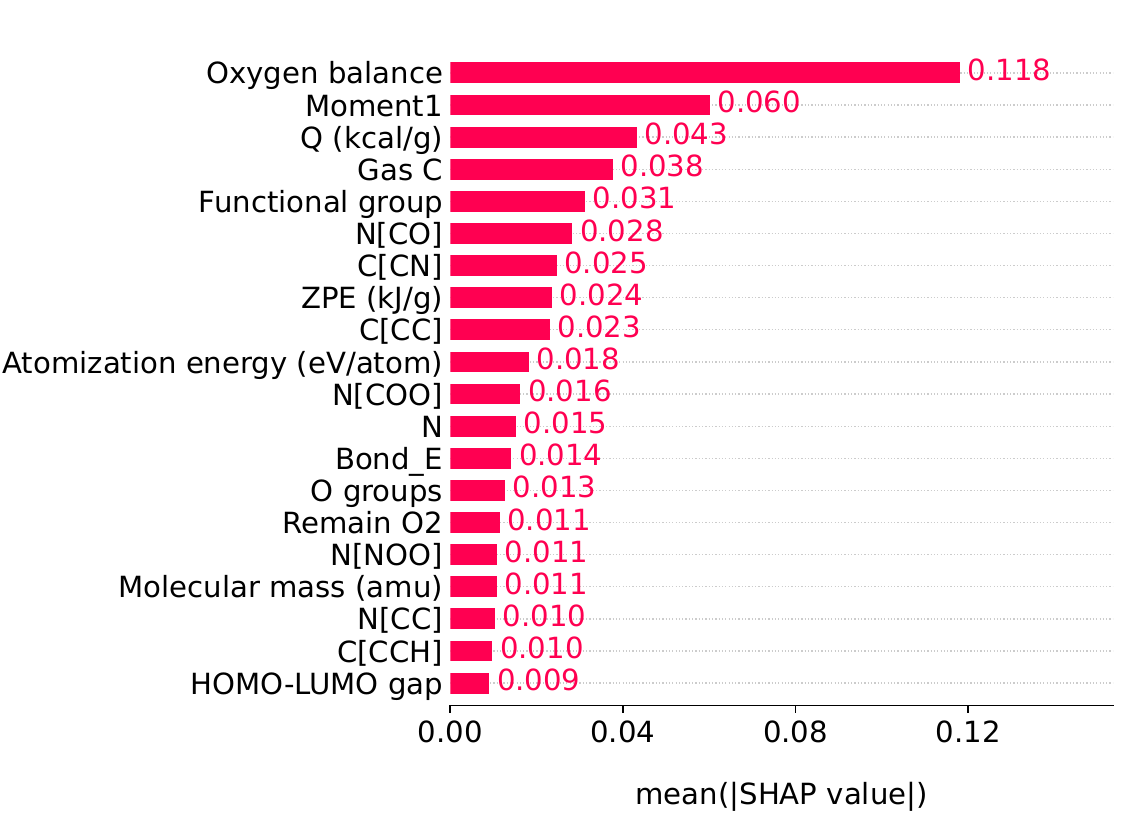}
    \end{subfigure}
    \hfill
    \begin{subfigure}{0.495\textwidth}
        \includegraphics[width=\linewidth]{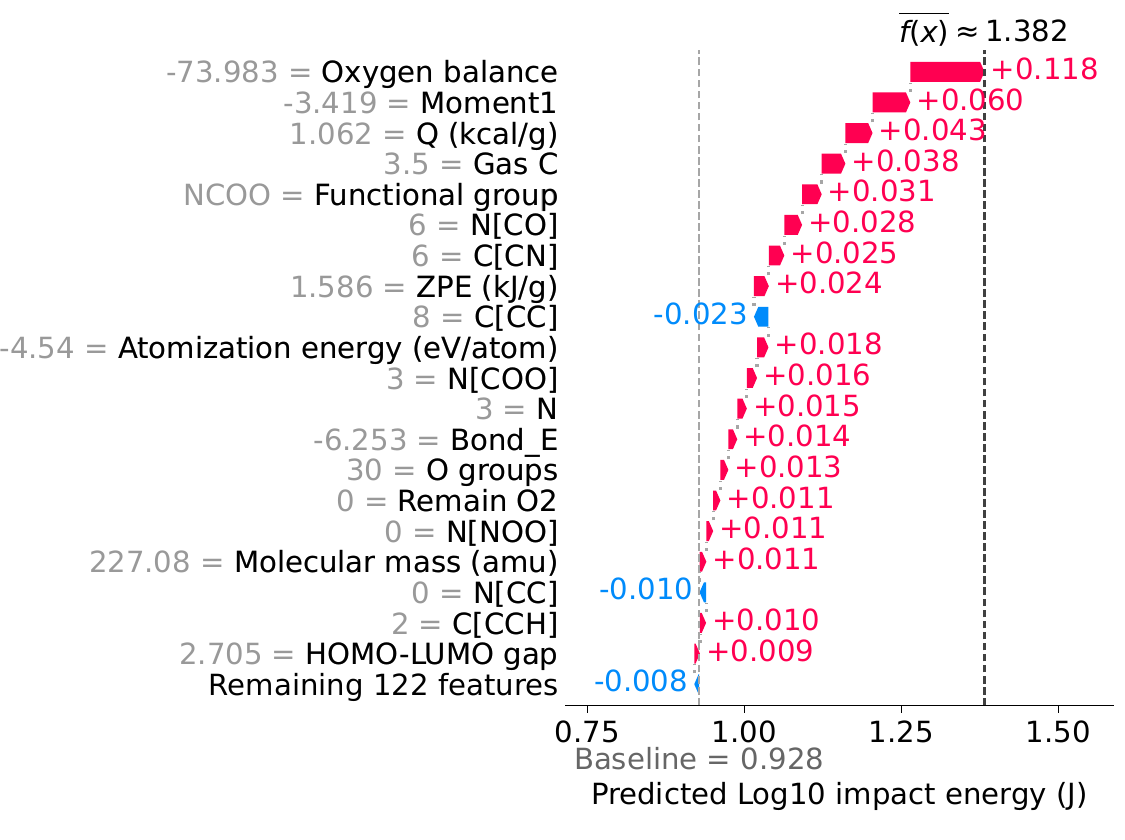}
    \end{subfigure}
    \caption{Shapley values for TNT, averaged over its 40 observations. The panels are as
    in Figure~\ref{fig:petn_shap}.}
    \label{fig:tnt_shap}
\end{figure}

Figures~\ref{fig:hmx_shap}--\ref{fig:tnt_shap} show the same summaries for HMX, RDX, NQ and
TNT, computed and plotted as for PETN in Figure~\ref{fig:petn_shap}.
The baselines differ between the compounds because each compound's TabPFN has its own context.
For HMX and RDX, the average prediction lies below the baseline, as for PETN.
For NQ and TNT, it lies above, so the model predicts these two compounds to be less
sensitive than the average observation in their context.
Oxygen balance and Moment1 are among the ten most important descriptors for all five
compounds. Both push the prediction towards higher sensitivity for PETN, HMX and RDX, and towards lower
sensitivity for NQ and TNT.

\clearpage

\bibliographystyle{plainnat-modified}
\bibliography{references}

\end{document}